# Sustainable LLM Inference using Context-Aware Model Switching

Yuvarani [a], Akashdeep Singh [a], Zahra Fathanah [a], Salsabila Harlen [a], Syeikha Syafura Al-Zahra binti Zahari [a], Hema Subramaniam [a]

[a]*Faculty of Computer Science and Information Technology, Universiti Malaya, 50603 Kuala Lumpur, Malaysia*

**Abstract**

Large language models have become central to many AI applications, but their growing energy consumption raises serious sustainability concerns. A key limitation in current AI deployments is the reliance on a one-size-fits-all inference strategy where most systems route every request to the same large model, regardless of task complexity, leading to substantial and unnecessary energy waste. To address this issue, we propose a context-aware model switching approach that dynamically selects an appropriate language model based on query complexity. The proposed system uses a Context-Aware Model Switching for Energy-Efficient LLM Inference that combines caching for repeated queries, rule-based complexity scoring for fast and explainable decisions, machine learning classification to capture semantic intent, and a user-adaptive component that learns from interaction patterns over time. The proposed architecture was evaluated using real conversation workloads and three open-source language models (Gemma3 1B, Gemma3 4B and Qwen3 4B) with different computational costs, measuring energy consumption (via NVML GPU power telemetry), response latency, routing accuracy, and output quality (BERTScore F1) to reflect real-world usage conditions. Experimental results show that the model switching approach can reduce energy consumption by up to 67.5% compared to always using the largest model while maintaining a response quality of 93.6%. In addition, the response time for simple queries also improved significantly by approximately 68%. These results show that model switching inference offers a practical and scalable path toward more energy-efficient and sustainable AI systems, demonstrating that significant efficiency gains can be achieved without major sacrifices in response quality.

## 1. Introduction

The rapid growth of large language models (LLMs) has significantly changed how artificial intelligence is used today. Models such as GPT-4, Claude, and LLaMA demonstrate impressive capabilities in understanding and generating human language, enabling applications ranging from software development and content creation to decision support. As AI systems scale to millions of daily interactions, the environmental costs of implementing these models are becoming a growing concern. Training a single large language model can produce carbon emissions comparable to the lifetime emissions of multiple vehicles [1]. More importantly, inference, the repeated use of models after training, accounts for most of their total energy consumption over time, yet this aspect is often overlooked in discussions on AI sustainability [2]. Some studies have also explored model cascade and routing strategies [3].

A major cause of inefficiency lies in how AI systems are currently implemented. Most real-world systems use a single inference strategy, meaning all user requests are handled by the same large model, no matter how hard the task is. This uniform allocation of resources leads to unnecessary computation, as simple requests such as greetings consume the same computational resources as complex tasks such as code generation. While foundational work by Strubell et al. [1] provides quantitative assessments of training costs and Schwartz et al. [2] introduced the conceptual distinction between "Red AI" and "Green AI", there remains a significant gap in frameworks that prioritize long-term inference efficiency. Recent research has explored model cascading and routing strategies as potential solutions. Chen et al. [3] demonstrated that cascade-based routing can match GPT-4 performance while reducing cost by up to 98%, while Ong et al. [4] showed that learned routing with preference data can achieve over two-fold cost reductions while maintaining response quality.

Despite these advances, current routing approaches face limitations. Cascade-based systems such as FrugalGPT [3] introduce additional latency due to their sequential query structure, as each model in the pipeline must produce and evaluate a response before the next is invoked. Single-pass routers such as RouteLLM [4] address this latency concern but rely on preference data that may not transfer across domain shifts. More broadly, many existing approaches lack empirical validation in production environments on locally hosted, open-source models. There remains a gap in accessible and scalable solutions that enable dynamic model switching without excessive hardware cost or architectural overhead. When multiplied by millions of daily requests, the inefficiency of uniform large-model deployment has a substantial environmental and economic impact.

In the present study, a context-aware model switching system for large language models is designed and implemented to address this problem directly. This study examines whether AI systems can switch between model sizes to reduce energy consumption without degrading response quality, identifies routing mechanisms that achieve an optimal balance between efficiency and accuracy, and estimates the realistic level of energy savings when handling the diverse mix of queries that characterize real-world usage. The experimental evaluation demonstrates that model switching can reduce energy consumption by up to 67.5% compared to always using the largest model, while maintaining a response quality of 93.6% and improving response latency by approximately 68% for simple queries.

There are several contributions this study makes. To the best of the authors' knowledge, this is the first study to demonstrate that a multi-level hybrid routing approach combining deterministic and learned classification strategies can significantly reduce inference energy consumption without compromising output quality in a fully local, open-source deployment. Second, it presents a practical system architecture that can be adapted for other AI applications, structured around modular components that support separation of concerns, controlled execution, and extensibility. Third, it provides empirical insights into the trade-offs between energy efficiency, response quality, and latency in model switching systems.

## 2. Related Work

The environmental costs of large-scale AI have been a concern since at least Strubell et al. [1] provided the first systematic quantification of energy consumption during NLP model training, demonstrating that training a single large transformer model with neural architecture search can produce carbon emissions comparable to the lifetime emissions of five automobiles. Their work was foundational in shifting community attention from benchmark performance toward computational cost as a legitimate evaluation criterion. Schwartz et al. [2] built on this framing through the Green AI framework, formally distinguishing between "Red AI"—efficiency-indifferent research driven by scale—and "Green AI", which treats computational efficiency as a core objective alongside accuracy. Their proposal of floating point operations per second (FLOPs) as a universal efficiency metric has since influenced how researchers report the costs of model development. More recently, Luccioni et al. [8] extended this life-cycle perspective to the deployment phase by quantifying the carbon footprint of BLOOM, a 176-billion parameter open-access language model, across both training and inference. Their finding that inference emissions can rival or exceed training costs over a model's operational lifetime is particularly relevant to the present work, which targets inference efficiency specifically. Complementing these studies, Patterson et al. [5] conducted a large-scale empirical analysis of carbon emissions across a wide range of neural network training workloads, demonstrating that architectural and hardware choices—such as model sparsity and efficient accelerators—can reduce emissions by up to 1000× compared to naïve dense training approaches. Their findings reinforce the importance of system-level design decisions in minimizing the environmental impact of AI, a principle that extends naturally to the inference-time routing strategies explored in the present work.

A parallel body of work has addressed the inefficiency of deploying a single large model to all queries regardless of complexity. Chen et al. [3] introduced FrugalGPT, a cascade-based framework that routes queries through a sequence of increasingly capable models and halts once a response of sufficient reliability is produced. Their evaluation demonstrated that this strategy can match GPT-4 performance with cost reductions of up to 98%, establishing the technical feasibility of heterogeneous model deployment. However, cascade approaches introduce additional latency due to the sequential query structure, as each model in the pipeline must produce and evaluate a response before the next is invoked. Ong et al. [4] addressed this limitation through RouteLLM, a routing framework trained on human preference data from Chatbot Arena that routes each query to a single model rather than a sequential chain. By framing the routing decision as a binary classification problem between a strong and a weak model, RouteLLM achieves cost reductions of over two times on standard benchmarks while maintaining response quality, and its routers demonstrate meaningful transfer learning capabilities across unseen model pairs.

Model compression represents an alternative but complementary approach to inference efficiency. Gou et al. [6] provided a comprehensive survey of knowledge distillation techniques applied to large language models, examining how capabilities from large teacher models can be transferred to smaller student models through output alignment, rationale-based training, and in-context learning distillation. The survey distinguishes between white-box distillation—which leverages internal model states—and black-box distillation, which relies solely on model outputs, noting that the unprecedented scale of modern LLMs introduces challenges that earlier distillation frameworks were not designed to handle. Quantization and pruning offer further routes to model compression. A comprehensive survey by Zhu et al. [7] classified model compression techniques into quantization, pruning, knowledge distillation, and low-rank factorization, examining the trade-offs between compression ratio, hardware compatibility, and task performance across different approaches. A related approach to dynamic efficiency at the model level is early exit inference, wherein intermediate transformer layers produce predictions and the network halts early when confidence is sufficiently high, bypassing later layers entirely. Xin et al. [13] demonstrated this principle in DeeBERT, showing that up to 40% of BERT's computation can be skipped on simple inputs without significant accuracy loss. Separately, Fedus et al. [14] introduced the Switch Transformer, a sparse mixture-of-experts architecture in which each token is routed to a subset of specialized feed-forward layers rather than the full network, achieving substantial parameter scaling without proportional increases in computation. These static compression methods and architectural alternatives are distinct from the dynamic routing approach proposed in the present work, which achieves efficiency by selecting among fully formed models at inference time rather than modifying internal computation.

Liao et al. [9] provided an empirical characterization of how common inference efficiency optimization—including continuous batching, speculative decoding, and quantization—affect energy consumption across different GPU configurations and serving frameworks. Their study found that optimization designed to improve throughput do not uniformly reduce energy use, and that the relationship between computational efficiency and energy efficiency is neither linear nor consistent across deployment settings. This finding motivates a system-level rather than purely algorithmic approach to sustainable inference, reinforcing the premise of the present architecture. Carbon-aware scheduling has also attracted attention as a deployment-level intervention. Radovanović et al. [10] demonstrated that aligning computing workloads with periods of lower-carbon electricity supply in cloud data centers can significantly reduce operational carbon emissions, though this approach depends on workload flexibility and temporal delay tolerance that real-time inference does not always permit.

Ma et al. [11] introduced LLM-Pruner, a structured pruning framework for large language models that selectively removes non-critical coupled structures based on gradient information, enabling significant parameter reduction with minimal task-agnostic retraining. Their work demonstrates that model capacity can be substantially reduced while preserving general-purpose capabilities, which is relevant in resource-constrained settings where deploying full-size models is impractical. In this context, dynamic model selection becomes particularly relevant, as routing queries to the smallest viable model reduces both memory pressure and per-inference energy draw. The routing architecture proposed in the present work is architecturally agnostic to hardware platforms and can in principle be deployed in edge environments, though the present evaluation focuses on local single-host deployment.

Query complexity estimation, which underpins any adaptive routing approach, has received less systematic treatment in the literature than model compression or cascade design. Hu et al. [12] introduced RouterBench, a benchmark for evaluating multi-LLM routing systems across a standardized set of tasks and model pairs, finding that an oracle router—which always selects the optimal model in hindsight—can outperform the strongest single model while substantially reducing inference cost. Their results highlight the performance ceiling available to routing systems and motivate the development of more accurate query complexity classifiers. The sentence embedding approach employed in the present work at the third routing level draws on the all-MiniLM-L6-v2 transformer, which produces task-agnostic semantic representations suitable for cosine similarity comparison against predefined task vectors. This approach builds on the Sentence-BERT framework introduced by Reimers and Gurevych [15], which adapted the BERT architecture with siamese and triplet network structures to produce semantically meaningful fixed-length sentence embeddings computationally efficient enough for large-scale similarity search. The resulting embeddings provide a lightweight and generalizable mechanism for classifying queries that fall outside the coverage of deterministic rule sets, without requiring the overhead of a full generative inference pass.

In aggregate, the existing literature establishes a clear direction: inference-time efficiency depends not only on the properties of individual models but on how queries are matched to models. Model compression addresses this at the architecture level by reducing model capacity; cascade and routing frameworks address it at the decision level by selecting among available models. The present work contributes to the latter category by combining a caching layer, a deterministic rule-based classifier, and a learned semantic classifier into a unified multi-level routing pipeline evaluated on real conversational workloads. To the best of the authors' knowledge, this is the first work to demonstrate that such a hybrid routing approach, deployed entirely on a single local host using open-source models, can achieve greater than 67% reduction in inference energy consumption while retaining more than 93% response quality relative to a large-model baseline.

### 3. Methods
**System Architecture**

The proposed system adopts a three-level hybrid routing architecture designed to achieve high throughput while minimizing energy consumption. The architecture follows a fast-path-first principle, whereby query classification is attempted using the least computationally expensive mechanism before escalating more resource-intensive processes. This hierarchical approach enables the system to optimize inference efficiency without compromising response accuracy.

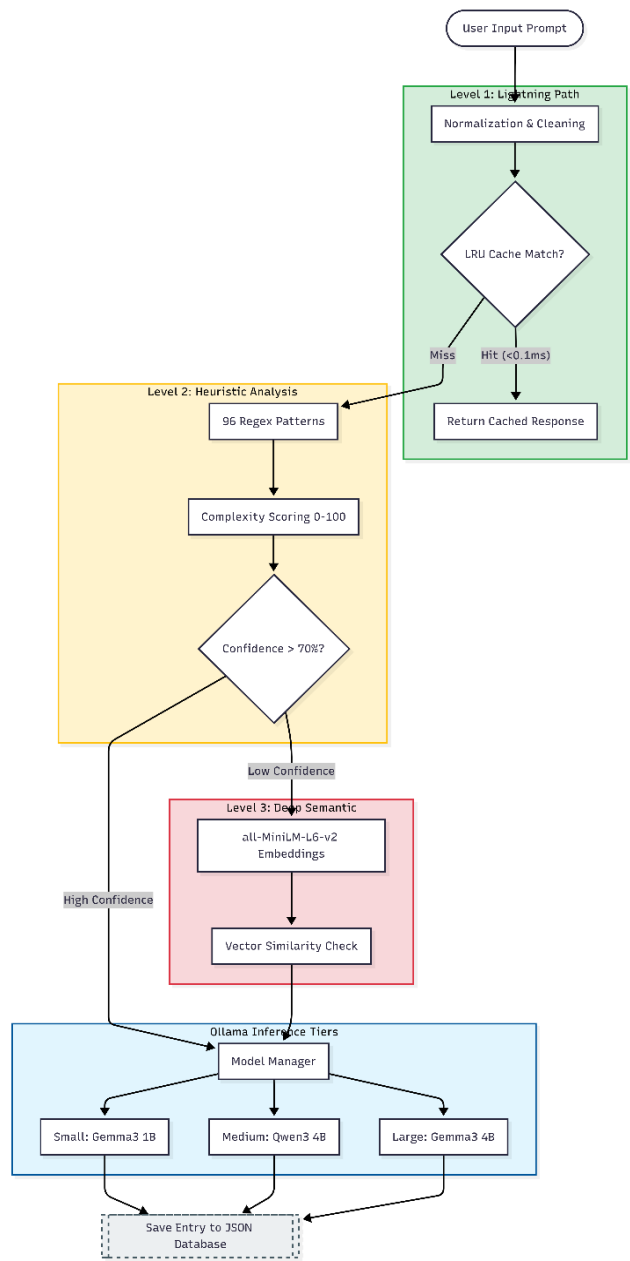

*Fig. 1. Multi-layer routing architecture.*

As illustrated in Fig. 1, the architecture is structured as a multi-layered, context-aware framework that balances computational precision and energy expenditure. The system is partitioned into three primary functional modules that operate as a synchronized processing pipeline: the Query Router, the Model Manager, and the Ollama API layer. Together, these modules enable adaptive model selection and efficient resource utilization based on query complexity.

The Query Router module functions as a multi-stage decision engine employing a waterfall routing strategy comprising cache lookup, heuristic rule evaluation, and machine learning–based semantic analysis. Each incoming query is assessed and assigned a normalized complexity score within the range of 0 to 100, which serves as the core criterion for downstream model selection. This staged routing mechanism ensures that low-complexity queries are resolved rapidly while reserving advanced inference resources for tasks that require deeper reasoning.

The Model Manager module is responsible for optimizing hardware resource utilization through dynamic model lifecycle management. To improve Video Random Access Memory (VRAM) efficiency and reduce idle power consumption, the system enforces a keep_alive: 0 policy. Under this configuration, only the actively selected model instance is loaded into the hardware accelerator at any given time, and model memory is immediately released upon completion of inference. This strategy significantly reduces memory footprint and energy waste associated with idle models.

The Ollama API layer serves as the inference backend and hosts a heterogeneous ensemble of locally deployed language models. The system incorporates three model tiers: Small (Gemma3 1B), Medium (Gemma3 4B), and Large (Qwen3 4B) each mapped to predefined complexity thresholds determined by the routing engine. This tiered deployment enables fine-grained control over computational cost while maintaining task-appropriate inference quality. A high-level representation of the system architecture is shown in Fig. 2.

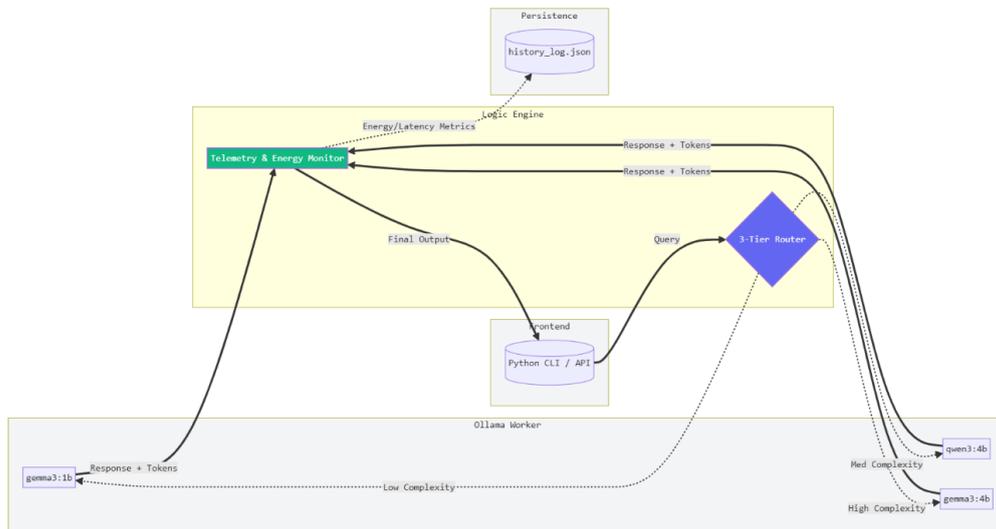

Fig. 2. High-level system diagram showing the three primary modules.

The routing efficiency of the system is primarily derived from its three-tier classification hierarchy, which is designed to resolve queries at the lowest feasible computational cost. At Level 1, an in-memory Least Recently Used (LRU) cache provides immediate resolution for frequently repeated queries. A Time-To-Live (TTL) of 300 seconds was selected as a pragmatic balance between cache freshness and hit rate, empirically determined during pilot testing to capture the majority of repeated queries within a conversational session. This layer achieves sub-millisecond latency, typically below 0.1 ms. Level 2 employs a deterministic pattern matching through a library of 96 precompiled regular expressions and keyword hash sets. This layer detects structural indicators such as programming syntax or mathematical operators and performs classification within a latency range of 0.1 to 1.0 ms. For queries that exceed the confidence threshold of Level 2, Level 3 invokes a semantic machine learning classifier. This classifier utilizes the all-MiniLM-L6-v2 transformer model, consisting of 22.7 million parameters, to generate sentence embeddings and compute cosine similarity against predefined task vectors.

Beyond these three routing levels, the system incorporates a user-adaptive component that refines routing decisions based on accumulated interaction history. This component maintains a lightweight per-session log of query–model pairings and observed quality outcomes. Over the course of a session, it adjusts complexity score thresholds incrementally when repeated misrouting patterns are detected. For example, if a user consistently submits queries of a particular domain that the rule-based layer underestimates. This adaptive mechanism does not modify the underlying model weights but instead tunes the routing boundary conditions at runtime, allowing the system to personalize inference allocation without retraining.

To support empirical evaluation, a dedicated Telemetry module was developed and integrated into the inference pipeline. This module intercepts the response generation process to measure real-time throughput in tokens per second and computes total energy consumption in Joules. These performance metrics are subsequently correlated with response quality to derive a sustainability index for the system, enabling quantitative analysis of efficiency–quality trade-offs.

The implementation stack was selected to ensure modularity, reproducibility, and experimental rigor. The system was implemented using Python 3.10 on Windows 11 (64-bit), leveraging its extensive support for asynchronous execution and machine learning libraries. Ollama was chosen as the inference orchestration engine due to its efficient handling of local model preloading and state management. The sentence-transformers framework was integrated to support semantic analysis at the third routing level. For data persistence, a schema-less JSON-based logging layer was implemented to record longitudinal measurements of energy consumption, latency, and routing accuracy for post-experimental validation.

**Methodology**

This study adopts a Design Science Research (DSR) methodology combined with experimental evaluation to address efficiency limitations in conventional large language model (LLM) inference systems. The primary objective is to design and validate a model switching architecture that mitigates latency, energy consumption, and carbon emissions associated with the prevalent one-size-fits-all inference approach. The proposed methodology is structured into four sequential phases, encompassing requirement analysis, architectural design, implementation, and comparative evaluation.

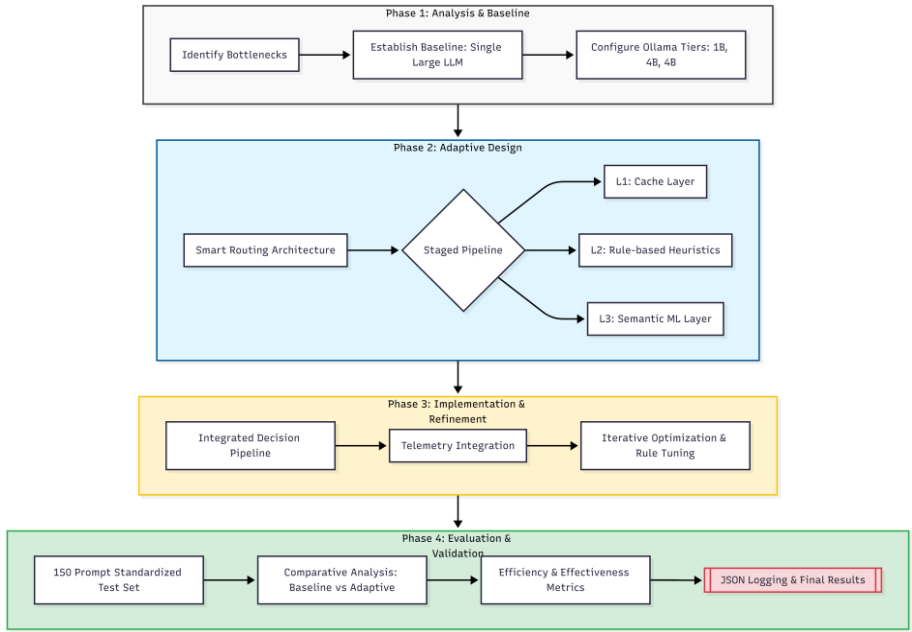

*Fig. 3. Four-phase research methodology*.

The first phase focuses on requirement analysis and baseline definition. In this phase, performance bottlenecks, excessive energy usage, and unnecessary carbon emissions in existing LLM deployment practices are identified. A baseline system is established in which all incoming queries are routed to a single large LLM that remains continuously loaded in memory, regardless of query complexity. To support adaptive inference, three tiers of locally hosted models are configured using the Ollama runtime: a small model (Gemma3 1B) for simple conversational tasks, a medium model (Gemma3 4B) for moderate reasoning workloads, and a large model (Qwen3 4B) for complex reasoning and code-related tasks. All experiments were conducted on a single local host machine running Windows 11 (64-bit) to eliminate network-induced variability and ensure controlled, consistent measurement conditions. The hardware configuration comprised an AMD Ryzen 7 5800H processor (8 cores, 16 threads, base clock 3.2 GHz), 32 GB DDR4 RAM, and an NVIDIA GeForce GTX 1650 Ti GPU (4 GB GDDR6 VRAM) with CUDA 11.8 support. GPU memory was shared with system RAM under the Windows unified memory model. Model inference was executed through Ollama with Python 3.10, leveraging PyTorch with CUDA acceleration where available. Energy consumption was measured using NVIDIA Management Library (NVML) GPU power telemetry combined with software-level timing via the Python time module, recording per-query energy in Joules as the product of mean GPU power draw (watts) and inference duration (seconds). CPU-side energy was not independently instrumented; reported figures reflect GPU-dominant inference energy. Carbon emissions were estimated using a global average carbon intensity of 475 $gCO_2e$/kWh as reported by the International Energy Agency [16].

The second phase involves the design of the model switching architecture, referred to as the Smart Routing Phase. This architecture introduces a staged decision pipeline in which computationally inexpensive mechanisms are applied prior to invoking costly model inference. Transitions between routing stages are governed by predefined confidence thresholds to preserve response quality. The routing strategy consists of three hierarchical levels: a cache-based layer (L1) that reuses prior routing decisions for repeated queries, a rule-based layer (L2) that applies deterministic lexical and structural heuristics to estimate query complexity, and a semantic machine learning layer (L3) that leverages sentence embeddings for cases where rule-based confidence is insufficient.

In the third phase, the proposed architecture is implemented and iteratively refined. The routing logic is realized as an integrated decision pipeline augmented with metrics collection and logging components. Initial experimental runs are used to identify misclassification patterns, which inform subsequent adjustments to routing rules and confidence thresholds. This iterative optimization process aims to improve routing accuracy while maintaining a balance between computational efficiency and output quality. In parallel, telemetry tools are incorporated to automatically capture key performance indicators, including end-to-end latency, throughput measured in tokens per second, and per-query energy consumption.

The final phase comprises comparative evaluation and validation of the proposed system against the baseline. A standardized evaluation dataset of 150 prompts was constructed to cover the distribution of queries typical in real-world conversational AI deployments. The dataset includes 50 simple queries (e.g., greetings, factual one-sentence questions, and common knowledge lookups), 50 medium-complexity queries requiring basic reasoning, multi-sentence explanation, or factual synthesis, and 50 complex queries involving multi-step reasoning, structured output generation, or code writing. Queries were manually curated to ensure categorical diversity and were reviewed for ambiguity before use. The dataset is available as supplementary material. The identical query set is executed on both the baseline system, which exclusively uses the large model, and the model switching inference. All experiments are conducted under identical hardware and runtime conditions, and results are averaged to reduce variance and improve reliability.

Evaluation metrics are categorized into efficiency and effectiveness measures. Efficiency metrics include end-to-end response latency, throughput, energy consumption per query, and estimated $CO_2$ emissions. Effectiveness is assessed through routing accuracy and output quality relative to the baseline system. Output quality is evaluated using BERTScore F1, a reference-based metric that computes token-level semantic similarity between candidate and reference responses using contextual BERT embeddings, which has been shown to correlate more reliably with human judgements than n-gram overlap metrics such as ROUGE. Responses from the adaptive system were compared against responses generated by the full large model baseline, with the baseline serving as the reference. While the limitations of automated evaluation methods are acknowledged, this approach enables consistent and scalable comparison. Experimental results demonstrate that the proposed smart switching system achieves significant reductions in latency and energy consumption, particularly for simple and medium-complexity queries, while maintaining high response quality for complex tasks. These findings confirm that adaptive inference can deliver substantial efficiency gains without significantly degrading user experience.

## 4. Results and discussion

The model switching inference underwent testing using a dataset comprising 150 queries distributed evenly across simple, medium, and complex categories. Each query was executed three times to minimize variance and ensure measurement stability. Automatically generated evaluation reports and execution logs provided the empirical foundation for the analysis presented here. Beyond standard evaluation protocols, the framework underwent stress testing with expanded query volumes to examine system stability, routing behavior patterns, and performance characteristics under sustained operational load. While quantitative analysis centers on the standard dataset, stress test observations contribute to qualitative insights regarding scalability and operational robustness.

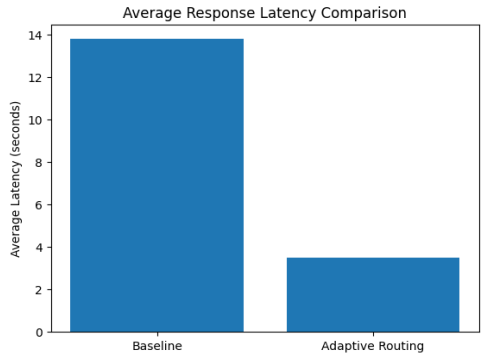

Fig. 4. Average response latency

Analysis of the standard evaluation reveals marked improvements in inference performance under adaptive routing conditions. The baseline system, which directs all queries to a single large model, demonstrated a mean response latency of 13.8 seconds per query ($\sigma = 2.1$ s). The adaptive system reduced this to a mean of 3.5 seconds ($\sigma = 1.4$ s), achieving a 68% reduction in end-to-end response time (Fig. 4). Throughput metrics exhibited parallel improvements. The adaptive configuration achieved an average generation rate of 61.3 tokens per second compared with 25.4 tokens per second under baseline conditions, representing a 141% throughput increase. These performance gains derive primarily from directing simple and intermediate queries to smaller, more efficient models, which reduces execution time and resource contention across the workload.

Energy consumption measurements revealed that model switching substantially diminishes total inference energy requirements. The baseline system consumed a mean of 84.2 kJ for the standard 150-query dataset ($\sigma = 6.3$ kJ across three repetitions), whereas the adaptive system required a mean of 22.0 kJ ($\sigma = 1.8$ kJ), yielding a 67.5% energy reduction (Fig. 5).

Energy consumption measurements revealed that model switching substantially diminishes total inference energy requirements. The baseline system consumed a mean of 84.2 kJ for the standard 150-query dataset ($\sigma = 6.3$ kJ across three repetitions), whereas the adaptive system required a mean of 22.0 kJ ($\sigma = 1.8$ kJ), yielding a 67.5% energy reduction (Fig. 5). This improvement manifests most dramatically for simple queries, where both latency and energy consumption declined to minimal levels through routing to the smallest model tier. Intermediate queries similarly benefited from reduced energy demands, while complex queries necessarily consumed greater energy through escalation to higher-capability models. However, even for complex queries, the adaptive approach eliminated unnecessary overhead by preventing redundant or misrouted executions.

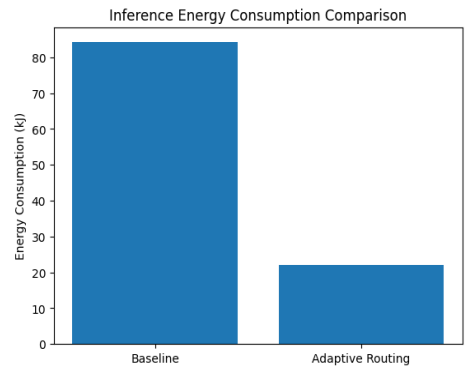

Fig. 5. Inference energy consumption

Applying a global average carbon intensity factor, the observed energy reduction translates directly into diminished carbon footprint. The baseline system generated approximately 11.1 gCO2e, while the adaptive system produced approximately 2.9 gCO2e, corresponding to a 67.5% emissions reduction. These outcomes demonstrate that inference-time optimization delivers measurable environmental benefits, especially in high-frequency deployment scenarios. Although absolute emission values vary with geographic location and energy source composition, the relative reduction remains consistent across contexts, confirming adaptive routing's viability as a sustainability mechanism.

Routing accuracy and classification performance received evaluation through precision, recall, and F1 metrics across query categories. The adaptive system attained overall routing accuracy of 79.3% with a weighted F1 score of 78.1% across all categories. Performance characteristics varied meaningfully by category. Simple queries achieved very high recall (98%), confirming the system's reliability in identifying low-complexity requests. Intermediate queries displayed balanced precision and recall, yielding robust F1 performance. Complex queries demonstrated high precision (96.3%) but comparatively lower recall (52%), reflecting the system's conservative escalation strategy. This behaviour deliberately prioritizes correctness and output quality over aggressive model downgrading, an appropriate design choice for production environments where errors on demanding tasks impose greater costs than efficiency losses.

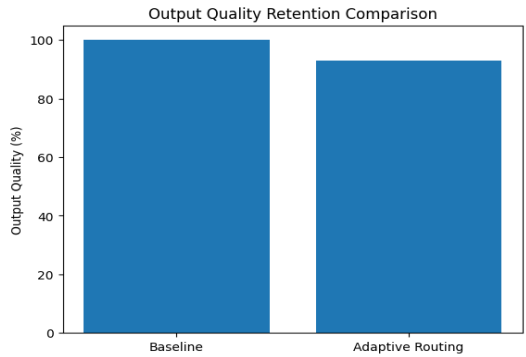

*Fig. 6. Output quality retention (BERTScore F1)*

Output quality assessment examined alignment between adaptive system responses and baseline large-model outputs using BERTScore F1. The adaptive system achieved a mean BERTScore F1 quality retention rate of 93.6% ($\sigma = 0.04$ across query categories), indicating that most responses closely approximated baseline outputs in both content and structure (Fig. 6). Minor quality degradation appeared principally in borderline cases where smaller models processed intermediate or complex queries. Conservative confidence thresholds limited such occurrences, and evaluation revealed no critical failures or unusable responses. These findings establish that model switching can yield considerable efficiency improvements while preserving acceptable response quality across most queries.

Category-level analysis exposed distinct behavioural patterns across query types. Simple queries derived maximum benefit, with average latency declining to approximately 300 milliseconds and energy consumption decreasing by more than an order of magnitude. Intermediate queries exhibited strong improvements in both latency and energy metrics, while complex queries showed moderate gains constrained by necessary model escalation. This differential performance confirms that the routing framework successfully allocates computational resources proportionate to task complexity rather than applying uniform processing across heterogeneous workloads.

Stress testing with expanded query volumes verified routing pipeline stability under elevated operational load. Cache hit rates increased with repeated queries, further diminishing average routing overhead. Execution sustained no routing failures or system crashes throughout stress test conditions. Although absolute latency grew under continuous load, the adaptive system-maintained performance advantages over baseline configuration, indicating favourable scaling characteristics. Comprehensive stress test logs remain available in evaluation artifacts for detailed examination. The user-adaptive component exhibited expected behaviour across conversational sessions. As session length increased, the component accumulated sufficient interaction history to begin adjusting complexity thresholds for domain-specific query patterns. In sessions involving repeated technical queries, the system progressively routed a higher proportion of borderline queries to the medium-tier model, reflecting learned calibration of the routing boundaries. While a full longitudinal evaluation of the adaptive mechanism falls outside the scope of the present study, these observations confirm that the component functions as designed and represents a productive direction for future empirical investigation.

The empirical evidence establishes that context-aware model switching achieves a favourable equilibrium between efficiency and quality. Performance and energy improvements stem predominantly from expedited resolution of simple and intermediate queries early in the routing pipeline, validating the architectural design principles articulated in the System Architecture section. Conservative handling of complex queries enhances system reliability and prevents unacceptable quality degradation, notwithstanding the resulting reduction in category-level recall metrics. From a deployment perspective, these findings indicate that adaptive model switching can materially reduce operational expenditures and environmental impact without necessitating model retraining or specialized hardware infrastructure. The approach demonstrates particular suitability for applications characterized by heterogeneous workloads and substantial query volumes.

Several constraints warrant acknowledgment. The evaluation concentrated on single-host deployment and did not examine effects of high concurrency or distributed architecture. Quality assessment relied on automated similarity metrics rather than human evaluation, potentially overlooking subtle semantic distinctions in responses. Furthermore, stress test results received qualitative rather than quantitative treatment in this analysis. Notwithstanding these limitations, the consistency of outcomes across performance, energy consumption, and quality metrics furnishes compelling evidence for the proposed approach's effectiveness. These constraints simultaneously indicate productive directions for subsequent investigation.

## 5. Conclusion

This study presents a context-aware, context-aware model switching system that addresses critical sustainability challenges in large language model inference. Through a multi-layer routing architecture combining caching, rule-based classification, and semantic machine learning, the proposed approach dynamically selects appropriately sized models based on query complexity. Empirical evaluation demonstrates that the system achieves a 67.5% reduction in energy consumption and 68% improvement in response latency while maintaining 93.6% response quality compared to baseline single-model inference. These gains stem from intelligent workload distribution rather than model modification, demonstrating that significant efficiency improvements are achievable through architectural innovation alone.

The key contributions include, to the best of the authors' knowledge, the first comprehensive demonstration on locally hosted open-source models that hybrid routing combining deterministic and learned classification strategies substantially reduces inference energy without compromising quality, a practical modular architecture supporting incremental adoption and extensibility, and empirical insights into efficiency-quality trade-offs in adaptive systems. For organizations deploying large-scale conversational AI, energy and cost savings compound to meaningful environmental and operational impact. The modular design enables incremental implementation and supports future enhancements including distributed deployment, domain-specific classifier training, and extension to other task types.

While limitations exist, including evaluation limited to conversational workloads and single-host deployment, the consistency of results across performance, energy, and quality metrics provides compelling evidence for the approach's effectiveness. As large language models become increasingly prevalent, smart deployment strategies that match computational resources to task requirements represent actionable solutions for sustainable AI development. This work establishes that sustainability and performance are complementary rather than competing objectives, offering both a validated system and methodological framework for advancing energy-efficient AI deployment in production environments.